\documentclass{article}
\usepackage{spconf}
\usepackage{times}
\usepackage{epsfig}
\usepackage{graphicx}
\usepackage{hyperref}
\usepackage{amsmath}
\usepackage{amssymb}
\usepackage{multirow}
\usepackage[dvipsnames]{xcolor}
\usepackage{threeparttable}

\usepackage{multirow}
\def\onedot{.}
\def\eg{\emph{e.g}\onedot} 
\def\ie{\emph{i.e}\onedot}

\def\etal{\emph{et al}\onedot}

\title{USING MULTIPLE LOSSES FOR ACCURATE FACIAL AGE ESTIMATION}
\name{Yi Zhou$^{\star}$, 
Heikki Huttunen${^\star}$,
Tapio Elomaa${^\star}$}
\address{$^{\star}$Department of Computing Sciences, Tampere University, Finland \\
        }

\begin{document}

\maketitle

\begin{abstract}
Age estimation is an essential challenge in computer vision.
With the advances of convolutional neural networks, the performance of age estimation has been dramatically improved.
Existing approaches usually treat age estimation as a classification problem.
However, the age labels are ambiguous, thus make the classification task difficult.
In this paper, we propose a simple yet effective approach for age estimation, which improves the performance compared to classification-based methods.
The method combines four classification losses and one regression loss representing different class granularities together, and we name it as Age-Granularity-Net.
We validate the Age-Granularity-Net framework on the CVPR Chalearn 2016 dataset, and extensive experiments show that the proposed approach can reduce the prediction error compared to any individual loss. The source code link is \url{https://github.com/yipersevere/age-estimation}.

\end{abstract}


\section{Introduction}
Age estimation has a broad spectrum of applications, including video surveillance, customer profiling and general facial analysis.
Age estimation is known for the difficulty of collecting sufficient training images with precise annotations.
Since the ground truth age can only be accessed by authorities, a popular alternative is to utilise appearance-based estimations by a group of human annotators.
However, such apparent age estimation also needs to address ambiguous labels resulting from subjective assessment.

Existing approaches for age estimation can be generally grouped into two categories: classification-based methods and regression-based methods.
For classification-based methods \cite{rothe2015dex, chen2017using}, age estimation is treated as a classification problem, \ie, each year is considered a class.
For regression-based methods \cite{niu2016ordinal, chen2013cumulative, yang2018ssr}, age estimation is treated as a regression task.
Alternatively, other methods may apply different mechanisms, \eg, label distribution learning \cite{geng2013facial, geng2016label}.

In this paper, we propose a novel framework named Age-Granularity-Net, which combines four classification losses and one regression loss for separate output branches from a shared backbone model---thus adding tolerance to label ambiguity.
The final loss is the sum of all individual losses for separate branches.

More specifically, the four classification losses consist of age labels representing different levels of granularity.
For example, in addition to one-year grid with 100 classes, we add output branches for lower granularities, \eg, 20 classes (with 5-years grid) and 10 classes (with one-decade grid).
Moreover, the ultimate granularity is represented by a regression branch, which uses the mean squared error as the loss function.
Note that the Age-Granularity-Net framework can be applied to several well-known neural networks, so it is more of a framework than a network type.

\section{Related work}
\label{sec:related_work}


\begin{figure*}[ht]
\begin{center}
\end{center}
    \includegraphics[scale=0.70]{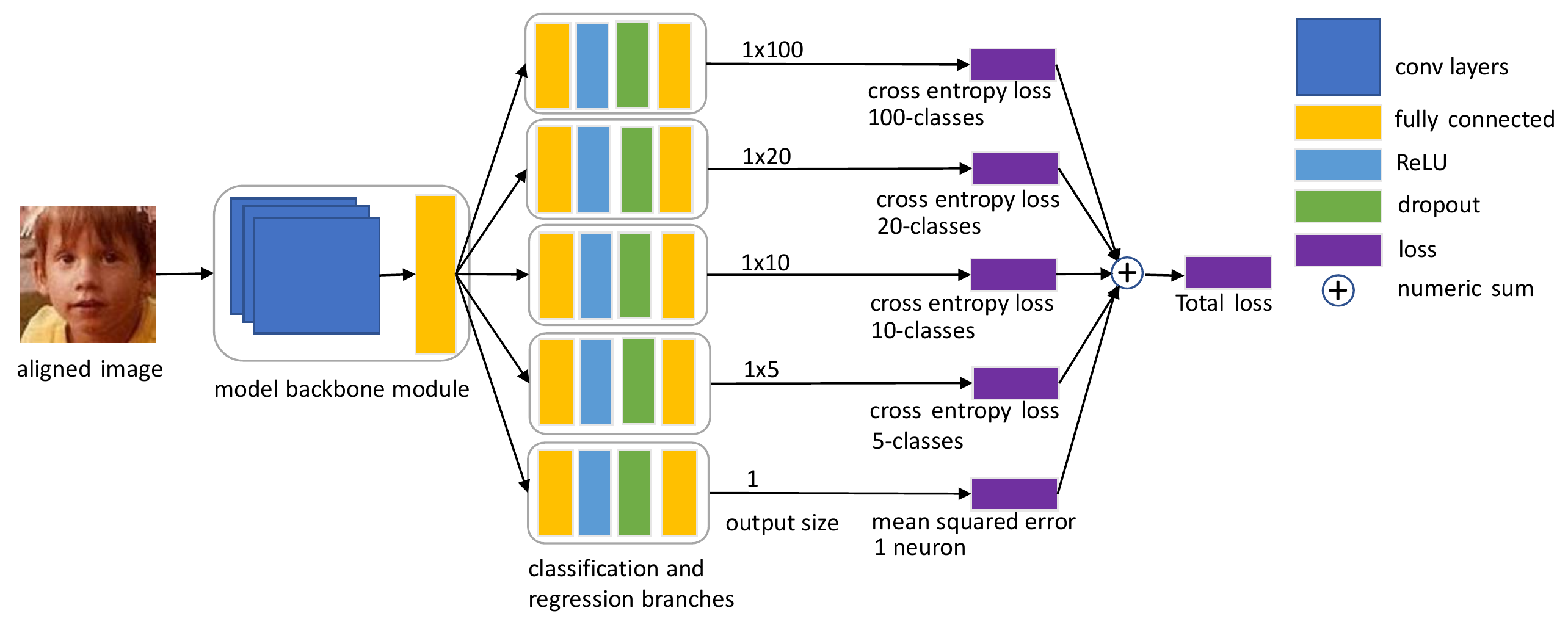}
   \caption{The architecture of the proposed method which combines different losses.}
\label{fig:architecture_cnn}
\end{figure*}

Age estimation is a representative task, where neural networks excel.
Until the surge of modern deep learning techniques, it was considered extremely challenging for computer programmes, while being quite straightforward for humans.
The root cause for this challenge is the holistic effect of ageing to the entire face, which is difficult to capture by classical machine learning methods.
However, convolutional neural networks can perceive higher-level concepts via their larger receptive field. Due to these reasons, recently CNNs have been adopted more widely than traditional approaches \cite{lowe1999object, ojala1994performance}.Yi \etal \cite{yi2014age} firstly utilized a CNN models to extract features from several facial regions. 

Classification and regression methods are two main approaches in age estimation. Regarding the classification methods for age estimation, Rothe \etal \cite{rothe2015dex} proposed to use the expected value of the probabilities on discrete age values as the predicted age, with each year of age representing one class.
Chen \etal \cite{chen2017using} utilized ranking-CNN for age estimation, in which there were a series of basic binary CNNs, aggregating to the final estimation.
Regarding the regression approaches for the age estimation, 
Niu \etal \cite{niu2016ordinal} formulated age estimation as an ordinal regression by employing multiple output CNNs.



Multiple losses approaches are used in age estimation task recently. Zhang \etal \cite{zhang2017quantifying} proposed to combine cross-entropy loss and distribution loss to quantify the facial age. Besides, Pan \etal \cite{pan2018mean} proposed the mean-variance loss as the final loss function, consisting of cross-entropy loss for the predicted age class and ground truth age class, L1-loss for the predicted age and ground truth age, and Kullback-Leibler divergence loss for the estimated age distribution and the probability distribution model of the ground truth age. 

Our approach differs from the related work in that we use multiple classification branches where each classifier has a task of different granularity constructed from the original labels.
The benefits of our approach are two-fold:
Firstly, using multiple granularities is a natural way to increase the tolerance against ambiguous labels---even nearby predictions get rewarded by the lower-granularity targets.
Secondly, using multiple simultaneous losses has been shown to improve the accuracy---possibly due to interpreting this as a form of label augmentation \cite{Bai2018, Zhou2019}.



\subsection{Attention models}

Much of progress in neural networks was enabled by so
called neural attention, which allows the network to focus
on certain elements of a sequence [xx], [xx], [xx] or certain
regions of an image [xx] when performing a prediction. The
appeal of such models comes from their end-to-end nature,
allowing the network to learn how to attend to or align data
before making a prediction.
It has been particularly popular in encoder-decoder frameworks, where it was first introduced to better translate between
languages [xx]. The attention network learned to focus on
particular words or phrases when translating sentences, showing large performance gains on especially long sequences. It
has also been extensively used for visual image and video
captioning, allowing the decoder module to focus on parts
of the image it was describing [xx]. Similarly, the neural
attention models have been used in visual question answering
tasks, helping the alignment between words in the question
and regions in the image [xx]. Spatial Transformer Networks
which focus on a particular area of image can also be seen as
a special case of attention [xx]. Somewhat relatedly, work in
facial expression analysis has explored using particular regions 
for facial action unit detection [xx], [xx], however, they did
not explore dynamically attending to regions depending on
the current facial appearance. Our work is inspired by these
attention models, but explores different ways of constructing
neural attention and applying it to age estimation.

\section{Proposed method}
\label{sec:proposed_method}

Due to ambiguity of age labels, we study the possibilities to increase the flexibility by introducing classification losses for targets at multiple granularities.
In the following, we consider four classification losses and one regression loss, but the number of losses is not limited to this configuration.

\subsection{Age-Granularity-Net Framework}

The neural network architecture of the proposed Age-Granula\-rity-Net framework is shown in Figure \ref{fig:architecture_cnn}. More specifically, we append five branches to any of the commonly used classification backbones. Each branch learns a variant of the age estimation task: The first branch estimates the age at 1-year granularity; the second one at 5-year granularity (20 classes; age target quantized to the mean of each 5-year interval); the third one at 10 year granularity and the fourth at 20 year granularity.
Additionally, we insert a regression output branch to represent the ultimately high granularity.

As mentioned, the proposed method can be easily applied to any neural network backbone.
In our experiments, we study the AlexNet \cite{krizhevsky2012imagenet}, MobileNet-V1  \cite{howard2017mobilenets}, ResNet50  \cite{he2016deep}, 
and DenseNet121 \cite{huang2019convolutional} models as the backbone.

\subsection{Aggregate Loss Function}

\begin{table*}[t]
    \begin{center}
      \scalebox{0.9}{
      \begin{threeparttable}   
        \begin{tabular}{|l|c|c|c|c|c|c|c|c|}
          \hline
          \multirow{2}{*}{Model}   & \multicolumn{4}{c|}{Classification loss combination}              & Regression   loss                                                     & \multicolumn{3}{c|}{Evaluation} \\ \cline{2-9}
                                                  & 100-classes & 20-classes & 10-classes & 5-classes  &  mse          & MAE           & improvement      & relative                            \\ \hline \hline
          \multirow{4}{*}{AlexNet}                & \checkmark  &            &            &            &            & 7.35          &                     &                         \\ \cline{2-9}
                                                  & \checkmark  & \checkmark &            &            &            & 6.81          & 0.54                &    7\%                     \\ \cline{2-9}
                                                  & \checkmark  & \checkmark & \checkmark &            &            & 6.86          & 0.49                &      7\%                   \\ \cline{2-9}
                                                  & \checkmark  & \checkmark & \checkmark & \checkmark &            & 6.11          & 1.24                 &       17\%                 \\  \cline{2-9}
                                                  & \checkmark  & \checkmark & \checkmark & \checkmark & \checkmark & \textbf{3.74\textsuperscript{1}}          &  3.61       &     49\%                                \\  \cline{2-8}   \hline \hline
          \multirow{4}{*}{MobileNet-V1}           & \checkmark  &            &            &            &            & 5.35          &                    &                          \\ \cline{2-9}
                                                  & \checkmark  & \checkmark &            &            &            & 4.90          & 0.45                &   8\%                      \\ \cline{2-9}
                                                  & \checkmark  & \checkmark & \checkmark &            &            & 4.71          & 0.64                &   12\%                        \\ \cline{2-9}
                                                  & \checkmark  & \checkmark & \checkmark & \checkmark &            & 4.63          & 0.72                 &   13\%                        \\ \cline{2-9}
                                                  & \checkmark  & \checkmark & \checkmark & \checkmark & \checkmark & \textbf{4.11}         &  1.24         &  23\%                                  \\ \cline{2-9}  \hline \hline
          \multirow{4}{*}{ResNet-50}              & \checkmark  &            &            &            &            & 4.84          &                       &                       \\ \cline{2-9}
                                                  & \checkmark  & \checkmark &            &            &            & 4.85          & 0.01                   &   0\%                   \\ \cline{2-9}
                                                  & \checkmark  & \checkmark & \checkmark &            &            & 4.59          & 0.25                   &  5\%                    \\ \cline{2-9}
                                                  & \checkmark  & \checkmark & \checkmark & \checkmark &            & 4.40          & 0.44                   &  9\%                    \\ \cline{2-9}
                                                  & \checkmark  & \checkmark & \checkmark & \checkmark & \checkmark & \textbf{4.03} & 0.81                    &   17\%                  \\ \cline{2-9}   \hline \hline  
          \multirow{4}{*}{DenseNet-121}           & \checkmark  &            &            &            &            & 6.16          &              &                                \\ \cline{2-9}
                                                  & \checkmark  & \checkmark &            &            &            & 5.48          & 0.68          &    11\%                           \\ \cline{2-9}
                                                  & \checkmark  & \checkmark & \checkmark &            &            & 5.34          & 0.82          &       13\%                        \\ \cline{2-9}
                                                  & \checkmark  & \checkmark & \checkmark & \checkmark &            & 5.16          & 1.00           &       16\%                          \\ \cline{2-9}
                                                  & \checkmark  & \checkmark & \checkmark & \checkmark & \checkmark & \textbf{3.86}  &   2.30         &       37\%                           \\ \cline{2-9}   \hline \hline
          \multirow{4}{*}{VGG-16-BN}              & \checkmark  &            &            &            &            & 5.35          &                 &                             \\ \cline{2-9}
                                                  & \checkmark  & \checkmark &            &            &            & 5.25          & 0.10             &    2\%                         \\ \cline{2-9}
                                                  & \checkmark  & \checkmark & \checkmark &            &            & 4.93          & 0.42          &       8\%                        \\ \cline{2-9}
                                                  & \checkmark  & \checkmark & \checkmark & \checkmark &            & 4.99          & 0.36           &      7\%                        \\ \cline{2-9}
                                                  & \checkmark  & \checkmark & \checkmark & \checkmark & \checkmark & \textbf{4.13} &  1.22          &      23\%                            \\ \cline{2-9} \hline \hline
          DLDL-v2 (ThinAgeNet)\textsuperscript{2} &             &            &            &            &            & 3.45         &                  &                            \\ \cline{2-9}  \hline 
        \end{tabular}
        \end{threeparttable}
      }
      \begin{tablenotes}
        \footnotesize
        \item 1 the initial learning rate is $10^{-5}$ instead of $10^{-3}$ when training the model in this experiment because of the convergence cross entropy issue.
        \item 2 the state-of-the-art accuracy on the CVPR 2016 ChaLearn dataset \cite{gao2018age}.
      \end{tablenotes}
    \end{center}
    \caption{The different granularity classification loss experimental age estimation results on the ChaLearn 2016 dataset. The lowest MAE is highlighted in bold.}
    \label{tab:the_combination_with_different_classification_loss}
\end{table*}

We define the aggregate loss function $\mathcal{L}$ as a combination of $N$ components: $N-1$ cross-entropy classification losses   $\mathcal{L}_1,\ldots,\mathcal{L}_{N-1}$ and one mean squared error $\mathcal{L}_\text{mse}$ regression loss. 
Also, we denote the ground truth for the $n$'th branch by $\mathbf{y}_n$  and the corresponding network outputs by
$\hat{\mathbf{y}}_n$ ($n=1,\ldots,N-1$); and the scalar ground truth and network output for the MSE loss by $y_N$ and $\hat{y}_N$, respectively.
In order to compensate for different magnitudes of the classification and regression loss, we define a weight $\lambda\in\mathbb{R}_+$, which compensates for the magnitudes of the two types of losses.

Formally, we can represent the aggregate loss as
\begin{equation}
    \label{Eq:total_loss}
    \mathcal{L} = \sum_{n=1}^{N-1}\mathcal{L}_n(\mathbf{y}_n, \hat{\mathbf{y}}_n)  + \lambda \mathcal{L}_\text{mse}({y}_N, \hat{{y}}_N).
\end{equation}
In our experiments, the weight $\lambda$ is set to 1, but might require tuning for regression problems with different range of values.


\textbf{Cross-entropy losses.}  In our experiments, we use four classification losses based on cross-entropy ($N=4$).
%
Thus, the aggregate loss is given by
\begin{align}\begin{split}
    {\cal L} &= \mathcal{L}_{100}(\mathbf{y}_{100}, \hat{\mathbf{y}}_{100}) + \mathcal{L}_{20}(\mathbf{y}_{20}, \hat{\mathbf{y}}_{20}) + \mathcal{L}_{10}(\mathbf{y}_{10}, \hat{\mathbf{y}}_{10}) 
    \\ 
    & + \mathcal{L}_{5}(\mathbf{y}_{5}, \hat{\mathbf{y}}_{5}) + \lambda\mathcal{L}_\text{mse}({y}_N, \hat{{y}}_N), 
    \end{split}
    \label{Eq:calculate_multiple_classification_loss}
\end{align}
where $\mathcal{L}_{100}$ represents the cross-entropy loss with 100 classes and 
 age granularity  one year, 
$\mathcal{L}_{20}$ represents the cross entropy loss with 20 classes and  age granularity  5 years, 
$\mathcal{L}_{10}$ represents the cross entropy loss and age granularity 10 years, and $\mathcal{L}_{5}$ represents the cross entropy loss with 5 classes and age granularity 20 years.

\begin{figure*}
	\includegraphics[width=\linewidth]{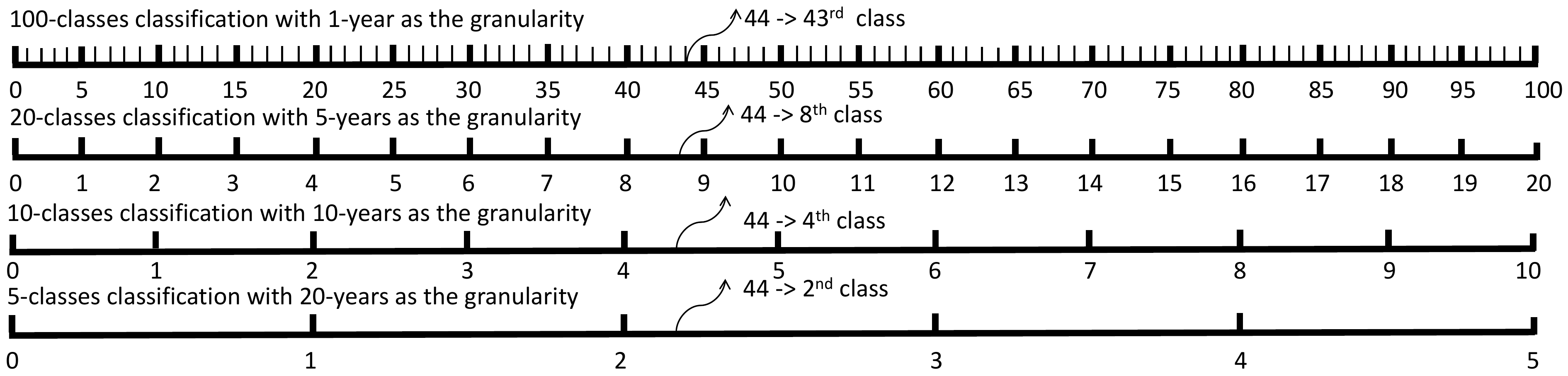}
	\caption{Illustration of the four classification branches on age estimation.}
	\label{fig:age_estimation_classification_loss}
\end{figure*}

The classification losses are illustrated in Figure \ref{fig:age_estimation_classification_loss}, where the ground truth of a sample image is 44 years. For the four classification losses, this is viewed as the ${43}$'rd class in the 100-class loss, the ${8}$'th class in the 20-class loss, the $4$'th class in the 10-class loss and ${2}$'nd class in the 5-class loss variant.



\textbf{MSE loss.} In addition to the $N-1$ classification losses utilizing cross-entropy, we also include another loss term defined by the mean squared error between the ground truth age and the network output from the regression branch produced by a dense single-output node with linear activation. 




\section{Experimental Setup}
\label{sec:experiments}


\subsection{Dataset}

In our experiments, we use the \textit{CVPR 2016 ChaLearn Looking at People dataset} \cite{escalera2016chalearn} with altogether 7,591 facial images (4,113 in the training set, 1,500 in the validation set and 1,978 in the test set) with human-annotated apparent ages. 
Images in this dataset were taken in non-controlled environments and had diverse backgrounds.






\subsection{Face landmark detection and alignment}

Face alignment process is a common procedure for age estimation.
We use the MTCNN alignment model \cite{zhang2016joint} to detect face and five facial keypoints: left eye, right eye, nose, left and right corner of mouth.
The detected face is aligned by similarity transformation obtained from the five keypoints.
For more details, refer to the alignment procedure of \cite{Bai2018}.



\begin{figure}
    \centering
	\includegraphics[width=0.9\linewidth]{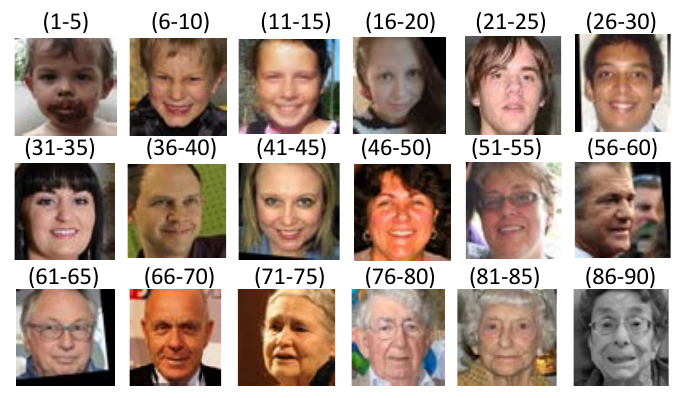}
	\caption{Example images with five years as the age granularity in CVPR ChaLearn 2016 dataset after face detection, alignment and data augmentation.}
	\label{examples_CVPR_ChaLearn_2016_dataset}
\end{figure}



\subsection{Preprocessing and Training}

Data augmentation is used in the experiments, including random crops with four paddings and a random horizontal flip. 
Examples of training samples are shown in Figure \ref{examples_CVPR_ChaLearn_2016_dataset} with five years as age granularity.
The ImageNet pre-trained weights are used to initialize the main backbone network, and the five branches are initialized at random.
All face images are resized to size 224 $\times$ 224 after alignment.

During the training phase, the initial learning rate is $10^{-3}$.
The learning rate is divided by 10 every time when the validation loss has not decreased for 8 epochs.
In the inference phase, we use the output from the ${\cal L}_{100}$ loss branch---with one year granularity.
This turned out to be the most accurate predictor among all five branches and their combinations.


\subsection{Evaluation metrics}
The mean absolute error (MAE) is the evaluation metric in our experiments. MAE is the average of absolute differences between the neural network predicted age and the ground truth age, where $k_{th}$ is one image from the $K$ total images.





\section{Experimental Results}

The age estimation accuracies using different numbers of losses are shown in Table \ref{tab:the_combination_with_different_classification_loss}. The table shows mean absolute errors in the test set when using five different backbone networks; each with five different combinations of loss functions. Moreover, we emphasize the difference to the single-loss version (with 100 classes, since this is the most common setup in the literature). 

From the table, it is clear that adding more losses increases the accuracy almost monotonously.
More specifically, almost every time when a new loss is added, the accuracy improves; suggesting that more losses are always better. One explanation for this that as we add losses with lower granularity, we also increase the tolerance to ambiguous labels.
The added accuracy brought by the new branches varies between 0\% and 11\% for the classification branches (5\% per branch on the average), while the relative decrease brought by the MSE loss branch ranges from 8\% for ResNet-50 to as high as 39 \% for the AlexNet (21\% on the average over all backbones).

Also, at the bottom is the state-of-the-art accuracy that a tailored architecture can reach on this dataset. Although the goal of this paper is not to compete against tailored age estimation networks, we can see that the reached accuracies are not far from those of the state-of-the-art.

\section{Conclusions}
\label{sec:conclusions}

In this paper, we proposed the Age-Granularity-Net framework for visual regression tasks. The method utilizes an aggregate loss function combined of losses at different target granularities. The key result is the observation that every added loss improves the accuracy, without increasing the computational load at inference: We only train with multiple losses, but infer with only one branch. Moreover, an important observation was that the MSE loss was very beneficial for the accuracy; indicating that the added losses should be complementary to each other in order to bring in novel aspects of the data.




\bibliographystyle{IEEEbib}
\bibliography{strings,yizhou_references}

\end{document}